%% file: main.tex
\documentclass{article}

\usepackage{neurips_2022}

\usepackage[english]{babel}

\usepackage{natbib}

\usepackage[utf8]{inputenc} 
\usepackage[T1]{fontenc}    
\usepackage{hyperref}       
\usepackage{booktabs}       
\usepackage{amsfonts}       
\usepackage{nicefrac}       
\usepackage{microtype}      
\usepackage{xcolor}         
\usepackage[pdftex]{graphicx}
\usepackage{subfig}
\usepackage{float}
\usepackage{mathtools}

\usepackage{amsmath}
\usepackage{graphicx}
\usepackage{float}
\usepackage{sidecap}
\usepackage{soul}
\usepackage{setspace}
\usepackage{authblk}

\sidecaptionvpos{figure}{t}

\usepackage[inline]{enumitem}

\graphicspath{ {./figs/} }

\DeclarePairedDelimiter{\norm}{\lVert}{\rVert}

\setcitestyle{numbers,open={[},close={]}}

\begin{document}
\title{A Penny for Your (visual) Thoughts: Self-Supervised Reconstruction of Natural Movies from Brain Activity} \singlespacing
\date{}
\input{authors.tex} \setstretch{0.5}

{\maketitlenew}
Project page: \href{https://www.wisdom.weizmann.ac.il/~vision/VideoReconstFromFMRI/}{\color{magenta}{\texttt{https://www.wisdom.weizmann.ac.il/$\sim$vision/VideoReconstFromFMRI/}}} \\



  \begin{SCfigure}[][h]
  \label{fig:Teaser}
    \label{fig:Teaser}
    \centering
    \subfloat{{
        \includegraphics[width=\textwidth]{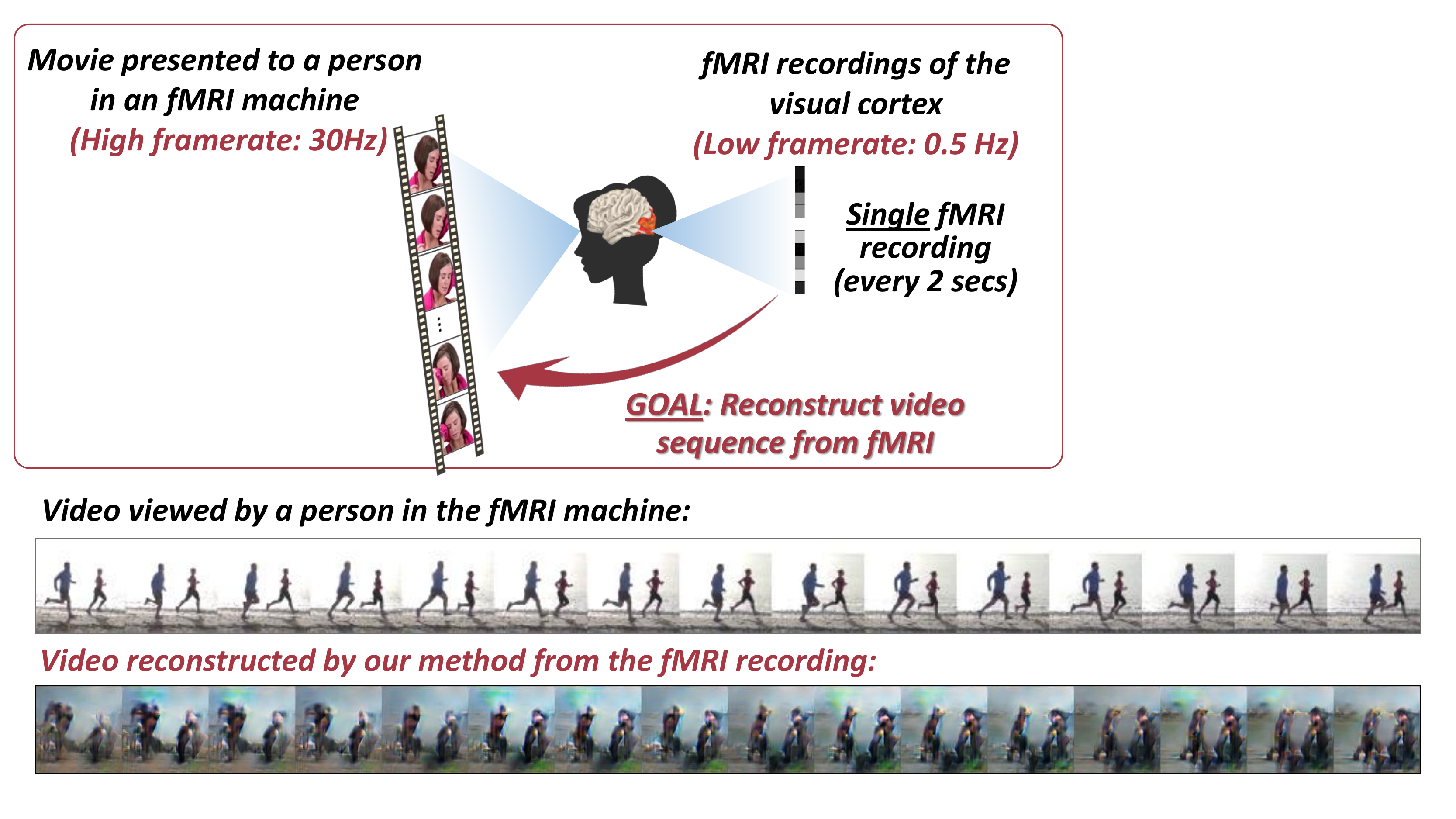} 
    }}%
   \hspace{-4cm}  
    \caption{\protect\rule{0ex}{4ex}\textbf{\mbox{\underline{Problem formulation:}} Natural movie reconstruction from fMRI brain recordings.}
    {\it We propose a self-supervised method for reconstruction of natural movies from brain activity, when only a limited number of paired training examples are available.
    }
    }
\end{SCfigure}

\vspace*{-0.3cm}
\def\thefootnote{\dag}\footnotetext{Equal contribution}\def\thefootnote{\arabic{footnote}}
\input{abstract.tex}

\clearpage
\input{introduction.tex}
\input{approach.tex}
\vspace*{-0.2cm}
\input{method.tex}
\vspace*{-0.2cm}
\input{results.tex}
\vspace*{-0.2cm}
\input{conclusion.tex}
\input{acknowledgments.tex}

 

    
    

\begin{ack}
\end{ack}

\medskip
\newpage
{
\small
\bibliography{references}

}

\clearpage
\newcommand{\beginsupplement}{%
        \setcounter{table}{0}
        \renewcommand{\thetable}{S\arabic{table}}%
        \setcounter{figure}{0}
        \renewcommand{\thefigure}{S\arabic{figure}}%
     }
\input{supplementary.tex}

\end{document}

%% file: authors.tex


\author{
 \textbf{ Ganit Kupershmidt$^\dag$, 
  Roman Beliy$^\dag$,
  Guy Gaziv,
  Michal Irani}
  }
\affil{Department of Computer Science and Applied Mathematics\\ The Weizmann Institute of Science \\ Rehovot, Israel \\}
\affil{\texttt{kganit94@gmail.com, \ \ roman.beliy@weizmann.ac.il}}

%% file: abstract.tex
\begin{abstract}
Reconstructing natural videos from fMRI brain recordings is very challenging, for two main reasons: (i)~As fMRI data acquisition is difficult, we only have a limited amount of supervised samples, which is not enough to cover the huge space of natural videos; and (ii)~The temporal resolution of fMRI recordings is much lower than the frame rate of natural videos. In this paper, we propose a self-supervised approach for natural-movie reconstruction. By employing cycle-consistency over Encoding-Decoding natural videos, we can: (i)~ exploit the full framerate of the training videos, and not be limited only to clips that correspond to fMRI recordings; (ii)~exploit massive amounts of external natural videos which the subjects never saw inside the fMRI machine. These enable increasing the applicable training data by several orders of magnitude, introducing natural video priors to the decoding network, as well as temporal coherence.  Our approach significantly outperforms competing methods, since those train only on the limited supervised data. We further introduce a new and simple temporal prior of natural videos, which – when folded into our fMRI decoder – further allows us to reconstruct videos at a higher frame-rate (HFR) of up to x8 of the original fMRI sample rate.

\end{abstract}

%% file: introduction.tex

\section{Introduction}
Decoding  natural visual stimuli from brain activity may be a cornerstone for brain-machine interfaces and for understanding visual processing in the brain. 
In particular, many works focused on the task of natural-image reconstruction from fMRI\footnote{functional magnetic resonance imaging.}-recorded brain activity \cite{Gaziv2022, Beliy2019, Cowen2014, Kamitani2005, Kay2009,Naselaris2009,Seeliger2018,Shen2019}. These works made strides in static stimulus reconstruction. However,  natural movies provide a closer, more realistic analog to dynamic visual experience~\cite{Nishimoto2011, Wen2018, Wang2022, Le2021}. The goal in this task is to
reconstruct unknown natural movies from a subject's fMRI brain recordings, taken while watching them (Fig~1).

Reconstructing natural movies from fMRI is significantly more challenging than natural-image reconstruction. Specifically, adding the temporal dimension gives rise to two \emph{additional} major challenges: 
(i)~There are not enough ``paired'' video-fMRI  training examples in current datasets  
to cover the huge pace of natural videos; and (ii)~The \emph{temporal} resolution of fMRI recording is extremely poor 
(an fMRI recording every 2secs = 0.5Hz),
in contrast to the very high video framerate
(30 frames/sec = 30Hz). 
Each fMRI sample is influenced by tens of video frames. In contrast, \emph{image}-fMRI datasets have a 1:1 correspondences between an fMRI sample and its underlying single image. 
Moreover, fMRI data are based on slow oxygenated blood flow. This signal's relation to the underlying brain activity is formally given by the Haemodynamic Response Function (HRF), which has a temporal support of more than a dozen seconds, with a peak after about 4 seconds. This results in a severe loss of high temporal frequencies.
It also poses additional challenges for recovering fast temporal information, or reconstructing movies with framerates higher than the fMRI sample-rate.

\textit{\textbf{Prior work on natural-movie reconstruction from fMRI}} can be broadly classified into two families: (i)~\emph{Frame-based methods}, which directly apply static \emph{image}-reconstruction methods to individual video frames in video-fMRI datasets~\cite{Wen2018,Han2019,Le2021}. These methods discard all valuable \emph{temporal} information across strongly correlated video frames, hence do not learn a model that captures 
space-time relations; (ii)~\emph{Video-based methods}, which aim to capture 
space-time relations by encoding/decoding small clips of a number of consecutive video frames~\cite{Nishimoto2011,Wang2022}.
This strategy allows to exploit temporal correlations within natural movies, and model their temporal dynamics. However, \cite{Nishimoto2011, Wang2022} are supervised methods, which train on the limited ``paired'' data alone. As such, they may be prone to overfitting the train-data and to poor generalization. 
Furthermore, temporal correlations exist also  across consecutive \emph{fMRI} samples. 
Ignoring the temporal coherence in the fMRI domain by decoding each fMRI sample independently,   results in abrupt, discontinuous transitions between reconstructed frames of successive fMRI samples.
To our best knowledge this was never addressed before. 

We present a new self-supervised approach for natural movie reconstruction from fMRI, which leverages natural video data well-beyond the few supervised fMRI-video ``paired'' examples. 
For this purpose, we train 2 separate networks: an Encoder (E) that maps a short video clip into an fMRI sample, and a Decoder (D) that maps fMRI samples to video frames. By concatenating these two networks back-to-back, Encoder-Decoder, we obtain a single network (E-D) whose input and output are video frames, hence its training does not require any fMRI recordings. We can thus vastly increase our training set to many ``unpaired'' data (natural videos without any fMRI). This increases our training data by several orders of magnitude.
Specifically, we apply this cycle-consistent self-supervised learning to train on: {(i)~Video clips from the provided training videos, but which do not have corresponding fMRI recordings ("Internal data"), and (ii)~Natural video clips from external video dataset ("External data")}. Applying this self-supervision to \emph{partially overlapping video clips}, 
not only dramatically expands the training data, but also teaches our decoder the notion of temporal coherence and temporal dynamics. Altogether, this results in a leap improvement in reconstruction quality over supervised-only training, and leads to  state-of-the-art reconstructions from fMRI.
%
%
We further developed a new self-supervised natural-video Temporal-Prior that enables video reconstruction at up to $\times$8 higher frame-rate (HFR) than the original fMRI sample-rate.



Adding self-supervision on unpaired data was previously proposed in the context of natural image reconstruction~\cite{Beliy2019,Gaziv2021,Gaziv2022}. However, their approach heavily relied on large \emph{external} natural-image datasets for self-supervised training. Our self-supervision for video reconstruction is significantly different. Firstly, unlike image-fMRI datasets, where there is full (1:1) image-fMRI correspondence, in video-fMRI datasets 
most frames have no corresponding fMRI
(video frames are at 30Hz; \ fMRI samples are at 0.5Hz). These provide a massive training data expansion by applying self-supervision \emph{internally}, giving rise to a leap improvement in reconstruction quality -- \emph{even without resorting to external datasets}.
Secondly, the consecutive frames in natural \emph{videos} are strongly correlated, providing additional valuable temporal information, which does not exist in natural \emph{image} reconstruction. 

Reconstructing natural videos at a higher frame-rate (HFR) than the fMRI's was recently proposed in~\cite{Wang2022}. We introduce a self-supervised temporal prior that, when incorporated into training, recovers substantially more temporal dynamics than~\cite{Wang2022}. Moreover, our self-supervision on ``unpaired'' data induces 
temporal continuity across reconstructions of \emph{successive fMRIs}, which~\cite{Wang2022} severely lacks.


 
\underline{Our contributions are therefore several-fold: }
\newline \ 
$\bullet$
A self-supervised approach to handle the inherent lack in  ``paired'' fMRI-movie training data.
\newline \ 
$\bullet$
State-of-the-art results in natural-movie reconstruction from fMRI.
\newline \ 
$\bullet$
\emph{Temporally-coherent} HFR video reconstruction -- higher than the underlying fMRI sampling rate.

%% file: approach.tex
\begin{figure}[t]%
\vspace*{-0.3cm}
    \centering
    \subfloat{\includegraphics[width=\textwidth]{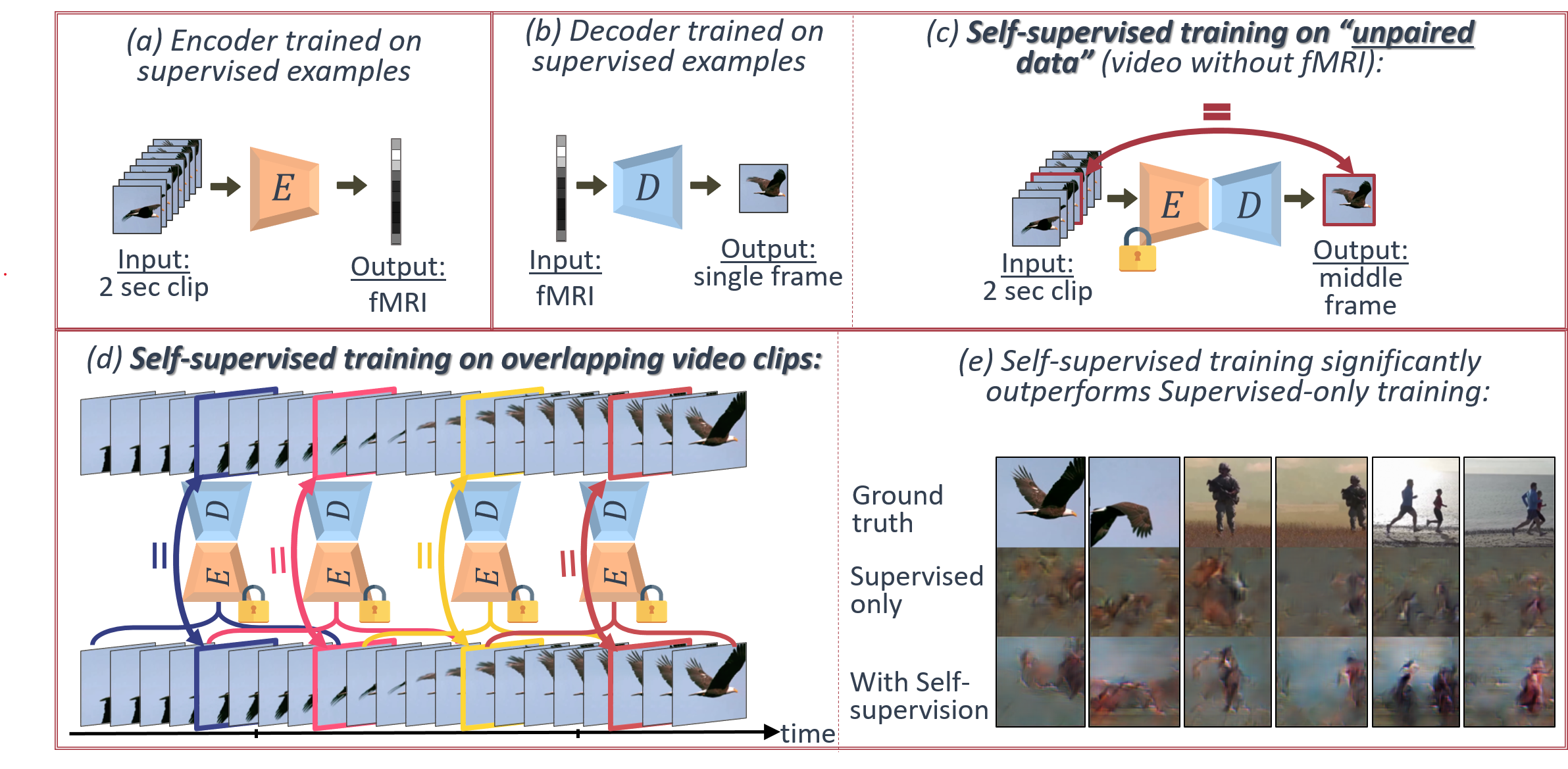}}
    \vspace*{-0.3cm}
    \caption{\textbf{Self supervised decoder training.} 
    {\it (a) We train a video-to-fMRI Encoder in a supervised manner. Then, we fix the pre-trained Encoder weights and train the Decoder using 2 types of examples: (b) Supervised training on ``paired'' examples, and (c) Cycle-consistent self-supervised training on ``unpaired'' videos clips, which have no fMRI recordings. (d) The cycle-consistent self supervised training exploits the full 30Hz framerate of the videos,increases the number of training data by several orders of magnitude, introduces natural video priors, as well as temporal coherence. (e) Adding self-supervision on ``unpaired data'' significantly outperforms supervised training only.}}
    \label{train fig}
\vspace*{-0.2cm}    
 \end{figure}

\section{Overview of the Approach} 
\label{approach sec}

\vspace*{-0.2cm}
We provide a high-level overview of the approach, whose components are later detailed in  Secs.~\ref{sec: self sup}, \ref{HFR}.

\vspace*{-0.2cm}
\subparagraph{Self-supervised video reconstruction from fMRI}
Our training consists of two phases. 
In the first phase, we train an Encoder model. It receives a 2-sec video and maps it to a single fMRI sample. It is trained in a supervised manner, using pairs of fMRI recordings and their corresponding video clips (see Fig.~\ref{train fig}a). In the second phase, we train a Decoder model with the help of 
the Encoder, whose weights are kept fixed.
The Decoder trains on two different types of samples: (i)~\emph{Supervised examples} (``paired'' video-fMRI examples)
and (ii)~\emph{Self-supervised examples} (``unpaired'' video clips with no fMRI).
Self-supervised training of the Decoder on ``unpaired'' video clips is made possible, by concatenating the 2 networks back-to-back, Encoder-Decoder (E-D), to a single network, with video frames as input and output.
In principle, video samples passed through the Encoder should \emph{ideally} predict an ``fMRI'', which when passed through the Decoder should \emph{ideally} return to the original input video.
Thus, E-D training does not require any fMRI recordings.
This self-supervised training allows to widely expand our training set, and train not only on supervised data (Fig.~\ref{train fig}b) but also on many ``unpaired'' video clips (Fig.~\ref{train fig}c).
Specifically, we apply this cycle-consistent self-supervised training on 2 types of ``unpaired'' data: (i)~Natural video clips from external video datasets ("External data"), and (ii)~Video clips from the video-fMRI
{training dataset}, but without corresponding fMRI recordings ("Internal data"). 
The latter allows to exploit all 30Hz overlapping clips in the training video, and not be restricted only to the 1/60 fewer ``paired'' clips. Why not to exploit this very rich temporal data?

Moreover, applying our cycle-consistent self-supervised training to \emph{partially overlapping video clips} (see Fig.~\ref{train fig}d), 
not only dramatically expands the available training data, but also teaches our Decoder the notion of temporal coherence and  dynamics. Altogether, this results in significant improvement compared to supervised-only training (see Fig.~\ref{train fig}e), and to  state-of-the-art reconstructions from fMRI.

\vspace*{-0.25cm}
\subparagraph{Higher frame-rate video reconstruction}
So far we examined the case of a Decoder that reconstructs video frames at the same temporal rate as the fMRI  (0.5Hz). We further extend our self-supervised approach to higher frame-rate (HFR) reconstruction. Our HFR Decoder receives two consecutive fMRI samples (see Fig.~\ref{hfr fig}a), and outputs multiple video frames (3 or more,  evenly spaced in the 2sec gap between the 2 fMRI samples). We train this decoder in the same manner as before, using both paired and unpaired examples (see Fig.~\ref{hfr fig}c), with an additional Temporal-Prior. 

\vspace*{-0.25cm}

\subparagraph{Incorporating a Temporal-Prior in the training.}
We request that our reconstructed frames not only be \emph{spatially} similar to their corresponding target frames, but also \emph{temporally} similar. A natural choice might be
similarity of optical flow. However, pre-trained optical-flow networks are not applicable here, since our reconstructed frames are very different than  (outside the distribution of) the video frames those optical-flow nets  trained on.
To overcome this, we learn an implicit temporal representation, by training a dedicated Temporal-Network on a simple self-supervised temporal task. It receives short video clips (5-9 frames) at 0.5-4~Hz. The frames are fed either in their correct order, or randomly shuffled. The model is trained to output whether the frames are shuffled or not (see Fig.~\ref{hfr fig}b). The model thus learns an implicit temporal representation of natural scene dynamics. We use the network's features from an intermediate temporal embedding, $T_{emb}$, and impose similarity of this embedding between the ground truth  and reconstructed frames, when training our decoder.

%% file: method.tex

\begin{figure}[t]%
\vspace*{-0.3cm}
    \centering
    \subfloat{{\includegraphics[width=\textwidth]{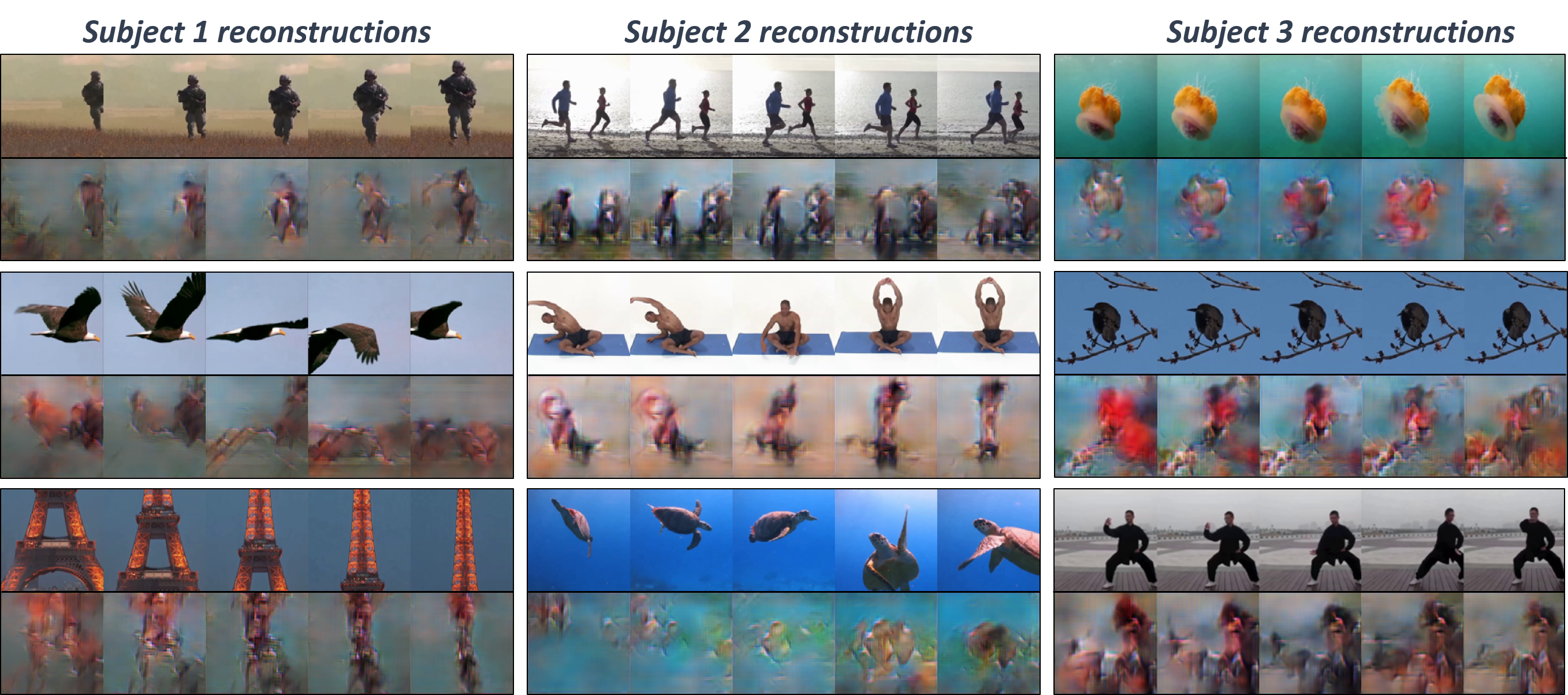} }}%
    \vspace{-0.2cm}
   \caption{\textbf{0.5 Hz reconstruction examples.} 
   {\it In each pair of rows: Top row shows the "ground  
   \mbox{truth" video frames; Bottom row shows our reconstructions from fMRIs of different subjects.}}}
    \label{decoder nice reconst}%
    \vspace*{-0.3cm}
\end{figure}

\section{The Algorithm}
\label{sec: self sup}
\subsection{Encoder training}
\label{Encoder method}
\vspace*{-0.2cm}
The Encoder model, E, encodes a 2 seconds video clip into a single fMRI sample of 5236 voxels\footnote{For more details on how those fMRI voxels were selected, as well as the SNR estimation of each fMRI voxel  -- see Appendix B.}
The input clip is sampled every other frame at a spatial resolution of 112x112. First, the Encoder transforms the short video clip into spatial-temporal features using the first two conv-layers of the pre-trained  action-recognition MARS network~\cite{Crasto2019}. 
This is followed by several separate temporal and spatial conv-layers, with a final FC layer to map the features to the fMRI space.
A figure showing the full architecture of the Encoder can be found in Appendix A
(Fig.~\ref{architect fig}a). 
While the MARS network  was trained with optical flow (OF) information (in addition to RGB frames), at test time it receives only RGB frames.
Thus, using the MARS pre-trained features allows us to exploit pre-trained dynamic information without requiring any OF. 
We train the Encoder using the following objective: 
\begin{equation}
    L_E(r,\hat{r}) = \norm{r- \hat{r}}_2 + \alpha \cos({\angle(r,\hat{r})})
\end{equation}

\vspace*{-0.2cm}
where $r, \hat{r}$ are the fMRI recording and its prediction, respectively, and $\alpha$ is an hyper-parameter (we used $\alpha=0.5$).
We started with initial learning rate of 1e-4, and reduce it every 3 epochs by a factor of $0.2$. We trained our Encoder for 10 epochs, on RTX8000 GPU, for a total training time of 20 minutes, and inference time of 13ms per video clip. 
Our Encoder is trained in a supervised manner only (using paired examples). Note that \emph{encoding} (going from multiple high-res frames into a single low-res fMRI) is a simpler task than \emph{decoding}.

\begin{SCfigure}
\hspace{-0.5cm}
    {\includegraphics[width=\textwidth]{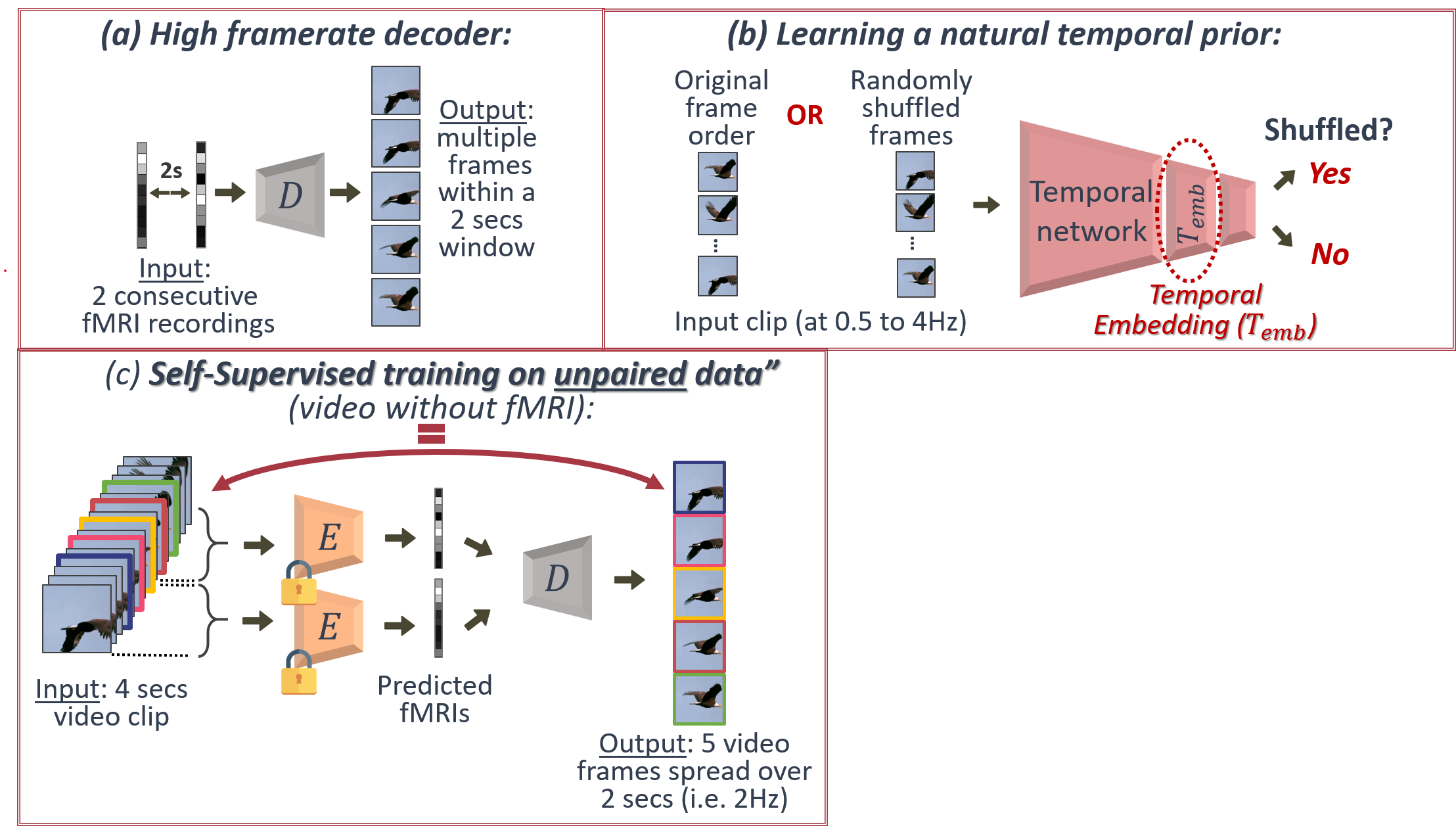}}
    \hspace{-6.5cm}
    \mbox{{\caption{\label{hfr fig}
\protect\rule{0ex}{23.5ex}\textbf{Higher frame-rate (HFR) video reconstruction.}
{\it  (a) Our HFR decoder receives 2 consecutive fMRI recordings, and reconstructs \emph{multiple} video frames in the 2-sec time gap. 
(b) We learn an implicit temporal embedding ($T_{emb}$) via  self-supervised training of a Temporal-Network. This net receives short video clips (5-9 frames), and outputs whether the frames are randomly shuffled or not  (see Sec.~\ref{temploss} for more details). (c) Self-supervised cycle consistency on ``unpaired'' video clips in the case of HFR decoder.}}}}
\vspace*{-1cm}
\end{SCfigure}

\subsection{Decoder training}
\vspace*{-0.2cm}
We begin with reconstructing video frames in the same temporal resolution as the fMRI (0.5Hz),
and with a spatial resolution of 112x112. For this purpose, our Decoder receives a single fMRI sample and outputs a single video frame. In each batch the Decoder, D,  is trained on 2 types of examples: supervised (paired) examples (Fig \ref{train fig}b), and 
self-supervised "unpaired" examples (video clips with no corresponding fMRI), trained using the concatenated network E-D.
At this stage E is fixed (Fig. \ref{train fig}c). 
Our Decoder is composed of a first fully connected (FC) layer
(to transform the input from the fMRI space to the frame space), followed by several alternating spatial conv-layers and upsampling.
A figure showing the full architecture of the Decoder  can be found in Appendix A
(Fig.~\ref{architect fig}b).


\textbf{The supervised Decoder loss} is composed of several components:
    \begin{equation}
    \label{eq:sigleframeloss}
         \mathcal{L}_{singleFrame}(x,\hat{x}) = \beta\mathcal{L}_{im}(x,\hat{x}) +  \gamma\mathcal{L}_{vid}(x,\hat{x}) + \delta\mathcal{L}_R(\hat{x})
     \end{equation} 
 where $x$, $\hat{x}$ are the ground truth frame and its reconstruction, respectively. The first component $\mathcal{L}_{im}$, is the perceptual similarity loss used in~\cite{Gaziv2022}, based on an intermediate spatial feature of the VGG image classification network~\cite{Simonyan2015}. The second component $\mathcal{L}_{vid}$ is a loss on features of the pre-trained P3D action recognition network~\cite{Qiu2017}, which can be applied to individual video frame. The $\mathcal{L}_{vid}$ loss minimizes the $l_2$ distance between the P3D-embeddings of the ground-truth frame and the reconstructed frame. Finally, $\mathcal{L}_R$ is a regularization term composed of: (i)~a total variation constraint to encourage smoothness of the reconstructed frames, and (ii)~a spatial group lasso regularization on the first FC layer (the layer which holds the majority of the network's parameters). 
 Finally, $\beta, \gamma, \delta$ are hyper-parameters (we used $\beta = 0.35, \gamma = 0.35, \delta=0.3$). 


\textbf{The self-supervised cycle-consistency loss:} 
The concatenated Encoder-Decoder network (E-D) receives as input a 2-sec video clip $v$, and is trained to output its middle video frame  $\hat{v}_{mid}$. The Encoder is fixed, and only the Decoder is trained.
The loss we use to impose cycle consistency on ``unpaired'' data is the same loss as in Eq.~(\ref{eq:sigleframeloss}). More specifically,
the objective is $\mathcal{L}_{singleFrame}({v}_{mid},\hat{v}_{mid})$, where $v$ is the 2-sec input clip, ${v}_{mid}$ is its middle frame, and $\hat{v}_{mid}$ is the reconstructed frame.

For the self-supervised training we used 2 types of ``unpaired data'': 
\textbf{``External data''} - natural videos that the subjects never saw, and \textbf{``Internal data''} - these are 2-second video clips within the 30Hz \emph{training videos}, but for which there are no corresponding fMRI recordings (as the fMRI is available only at 0.5Hz). The Internal data allows us to exploit the full temporal capacity of the training videos, and not only the sparse ``paired'' data (see Fig.~\ref{train fig}d). We  further apply augmentations to the unpaired clips (horizontal flips and temporal flips). 
Altogether, we \emph{increase the number of Internal training examples by 2 orders of magnitude} -- all from the \emph{same distribution} as the supervised ``paired'' data. Note that this does not require any additional external videos. Our experiments show that this type of ``unpaired data'' is the most significant part in our self-supervision (see Sec.~\ref{sec:experiments} and Table~\ref{table: ablation super}).

The Decoder was trained with \emph{equal weights} for the supervised loss and the self-supervised cycle-consistency loss. Each batch contained 64 ``paired'' examples and 64 ``unpaired'' examples. In addition, the model was trained with initial learning rate of 1e-3, which was reduced every 25 epochs by a factor of 5, over 100 epochs. The Decoder was trained on RTX8000 GPU, with a training time of 1.5 hours. The inference time is 0.5ms per fMRI recording.

  \begin{figure}[t]%
        \subfloat{{\includegraphics[width=\textwidth]{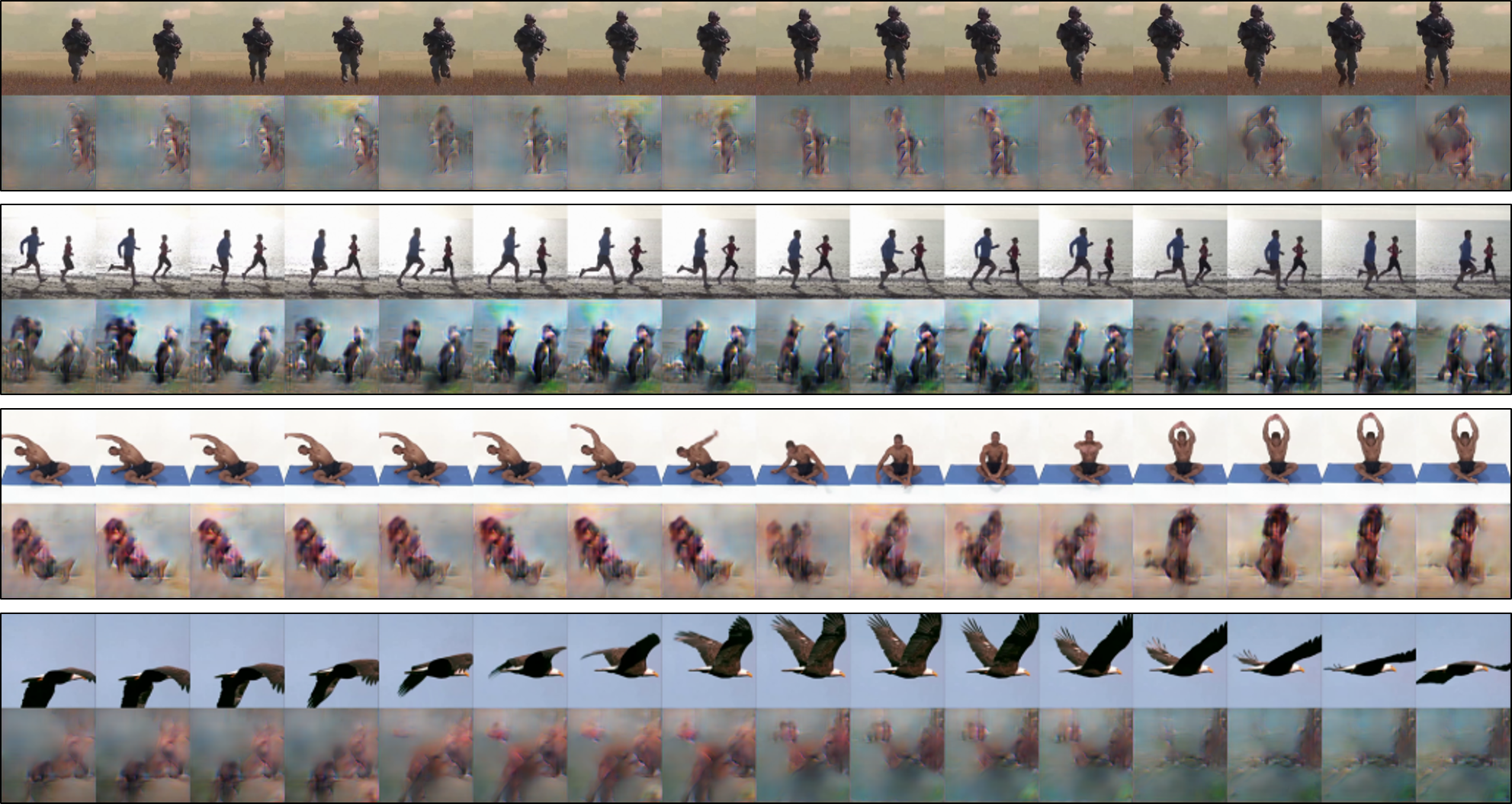} }}%
   \vspace*{-0.1cm}
        \caption{\mbox{\textbf{2 Hz reconstructions.} 
        {\it  In each pair: (Top) original frames; (Bottom) reconstructed frames}} 
        \label{decoder 1Hz fig}}
    \vspace*{-0.5cm}
    \end{figure}

\vspace*{-0.2cm}
\section{Recovering Videos at Higher Frame Rate (HFR)} 
\label{HFR}
\vspace*{-0.3cm}
In this section, we show how the Decoder network can be adapted to reconstruct video frames at higher framerate than the 0.5Hz fMRI sample-rate. Our higher framerate (HFR) Decoder receives 2 consecutive fMRI recordings, which allow the model to exploit also the temporal dynamics in fMRI space. The HFR Decoder learns to reconstructs multiple frames (3-9, depending on the desired framerate), evenly spaced across the 2-sec time gap between the 2 fMRIs. The first and last frame in the multi-frame reconstruction, are the middle-frame corresponding to each of the 2 fMRI input samples (same frames as in the 0.5Hz reconstruction); all other frames are extra  in-between frames.

We trained 3 types of HFR Decoders with different reconstruction rates: (i)~3 frames, 1-sec apart from each other (i.e. 1Hz reconstructions), (ii)~5 frames, 0.5-sec apart from each other (i.e. 2Hz reconstructions) and (ii)~9 frames, 0.25-sec apart from each other (i.e. 4Hz reconstructions). Since in this case we reconstruct more than one frame, we now also have a temporal dimension in the output space. Thus, in addition to the spatial appearance loss of Eq.~(\ref{eq:sigleframeloss}), we can now add also a \emph{temporal loss} on the reconstructed frames. 
To do this, we used the temporal embedding $T_{emb}$, learned by a self-supervised Temporal Network presented in Sec.~\ref{temploss}. We forward-feed both the ground truth clips and the reconstructed clips to the temporal-features space, and minimized the $l_1$ distance between their temporal embeddings $T_{emb}$.
The overall loss on our HFR Decoder is thus:
%
\vspace*{-0.1cm}
    \begin{equation}\label{eq:multiframeloss}
        \frac{1}{n}\mathcal{L}_{multiFrame} = \left(\sum_{i=1}^n \mathcal{L}_{singleFrame}(x_i,\hat{x}_i) \right) + \tau\mathcal{L}_{1}(T_{emb}(v),T_{emb}(\hat{v}))
    \end{equation}
    
\vspace*{-0.2cm}
    where $n$ is the number of reconstructed frames, $v$$=$$(x_1,x_2,..,x_n)$, $\hat{v}$$=$$(\hat{x}_1,\hat{x}_2,...,\hat{x}_n)$ 
    are the reconstructed videos (at the Decoder's reconstruction framerate). $\mathcal{L}_{singleFrame}$ is the loss from Eq.~(\ref{eq:sigleframeloss}, applied to each of the reconstructed frames.
    Since HFR reconstruction is a more ill-posed problem,
    we increased the weights of the regularization terms and the temporal term, at the expense of the spatial terms in Eq.~(\ref{eq:sigleframeloss}. Namely: $\beta$$=$$\gamma$$=$$0.25$, $\delta$$=$$0.35$, $\tau$$=$$0.15$.
Here too, each batch contains an equal number of ``paired'' and ``unpaired''  training examples, but unlike the 0.5Hz  decoder, the self-supervised loss has double weight compared to supervised loss.      
We used an initial learning rate of 5e-4, which was reduced every 25 epochs by a factor of 5, over 100 epochs. All HFR Decoders were trained on RTX8000 GPU, with training times of 3-12 hours. The inference time is 1.5-8ms. 
\subsection{Learning a Natural Temporal Video Prior}
\label{temploss}
We impose not only spatial constraints on our reconstructed video frames, but also temporal constraints. For that we use an intermediate representation of a Temporal Network, which was separately trained to identify if an input video clip is in its correct frame order, or in a shuffled frame order.
Note that such a network implicitly learns a notion of temporal dynamics in natural videos. Its training is simple and self-supervised, without any manual labels. 
Once trained, we use an intermediate representation of the network as a deep temporal embedding, $T_{emb}$, which serves as a temporal prior when training our HFR decoders (see Eq.~\ref{eq:multiframeloss}). 
We impose similarity between the temporal embedding of the sequence of reconstructed frames $T_{emb}(\hat{v})$, and the sequence of target frames $T_{emb}(v)$.

The inputs to our Temporal Network are short video clips of 5-9 video frames, temporally sampled at 0.5-4~Hz, at 64$\times$64 spatial resolution. 
The network is trained on the same video data  used to train our decoders (Internal \& External data). The loss is Binary cross entropy. Learning rate starts at 5e-4, and is reduced by a factor of 2 if the accuracy on a validation set does not improve. $T_{emb}$ is the outpu of the last convolution layer, of size (7,7,32).

%



The architecture of our Temporal Network is based on C3D~\cite{Tran2014}, and is as follows:
A 3D conv-layer, with 32 kernels of size (1,3,3), followed by 1$\times$2$\times$2 max pool. Then, another 3D conv-layer with 96 kernels of size (5,1,1), followed by temporal average pool. Next, a conv-layer with 32 kernels of size (3,3) with 2$\times$2 max pool, are applied twice. Finally, 2 FC layers with 512 output nodes each, and a last FC layer with a single output node.

%% file: results.tex
  \begin{figure}[t]%
        \centering
        \subfloat{{\includegraphics[width=\textwidth]{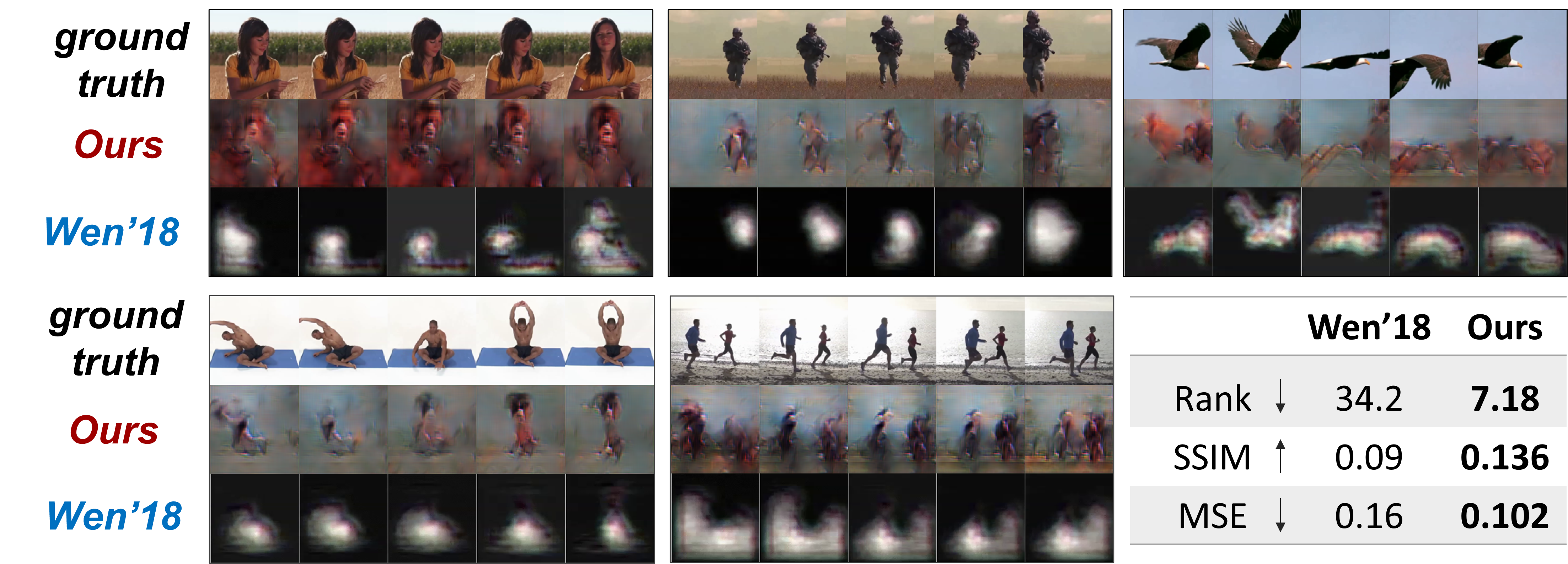} }}%
        \vspace{-0.2cm}
        \caption{\textbf{Comparison of 0.5Hz reconstructions.}
    {\it In every triplet of rows:
        (Top) the GT frames. (Middle) our reconstructions.
       (Bottom) reconstructions of~\cite{Wen2018}.
        The table shows quantitative comparison on the first 8-min test clip of Subject 1 (the only reconstructed clip provided by [Wen'18]).}}
             \label{SOTA visual comparison}%
       \vspace*{-0.4cm}
    \end{figure}
\clearpage
\section{Experiments and Results}
\label{sec:experiments}
\vspace*{-0.2cm}
\subsection{Experimental Datasets}
\vspace*{-0.2cm}
\textbf{fMRI dataset:} We used a publicly available benchmark fMRI dataset~\cite{Wen2018}, which contains fMRI recordings paired with their corresponding video clips. 
A 0.5Hz 3T MRI machine was used to collect fMRI data (with 3.5mm spatial resolution). 
There are recordings from 3 subjects (Subjects~1,2,3).  The training data comprises 18 segments of 8-minute video clip each -- a total of 2.4 video hours, which induce only 4320 corresponding fMRI frame recordings -- i.e., \emph{\textbf{4320 supervised ``paired'' training examples}}. The test data consists of 5 segments of 8-minute video clip each, resulting in 40 minutes of test-video -- i.e., \emph{\textbf{1200 test fMRIs}}.
The movies are very erratic, switching between short \emph{unrelated} scenes every 3-5 sec. Thus many fMRI recordings are affected by 2 different unrelated video clips (making
\mbox{the decoding problem even more difficult). All video clips have a temporal resolution of 30Hz.}

\begin{figure}
    \centering
    \subfloat{{
     \hspace*{0.1cm}   \includegraphics[width=\textwidth]{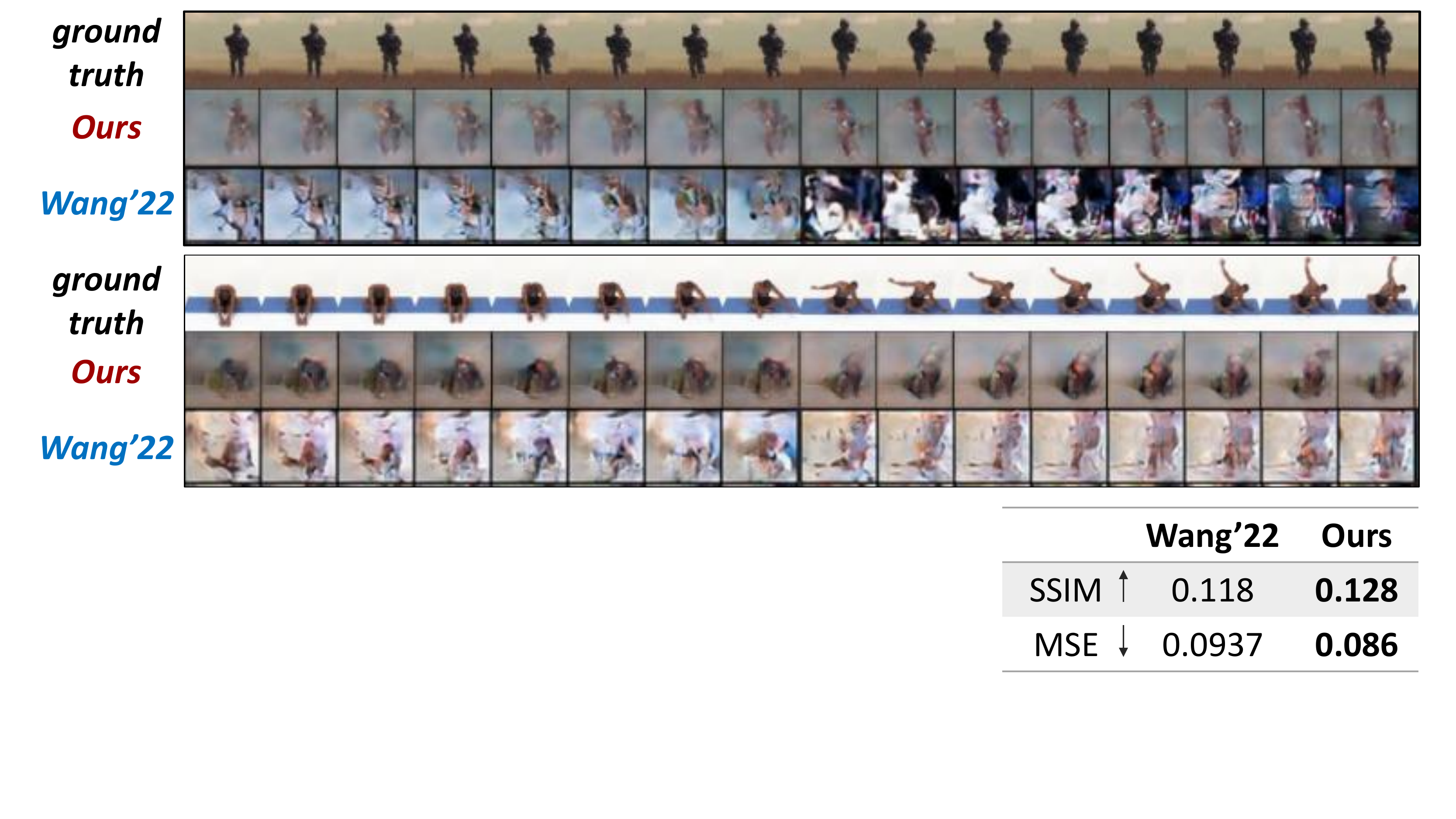} 
    }}%
    \vspace{-3.3cm} 
   \caption{\textbf{HFR Decoder (4Hz).} {\it (Top) visual comparison between\newline the original and HFR reconstructed frames. The first row shows the\newline GT frames, the second row shows our reconstructions and the third \newline row show Wang et. al. (2022) reconstructions. (Bottom-right) image\newline similarity comparisons, averaged over all test clips and  all subjects.}}
    \label{TSR SOTA visual comparison}
\vspace*{-0.5cm}
\end{figure}

\textbf{External video dataset (unpaired videos):} The external video dataset we used as part of  our self-supservised training was downloaded from \href{https://www.storyblocks.com/video}{\color{magenta}{\texttt{https://www.storyblocks.com/video}}}. These are 2.2 hours of natural videos, which were not viewed by the subjects, and have no fMRI recordings.

\vspace*{-0.3cm}
\subsection{Visual Results}
\vspace*{-0.2cm}
Fig.~\ref{train fig}e shows the huge difference in visual quality between supervised training the Decoder on ``paired'' examples only (fMRI with corresponding clips), versus adding also self-supervision on a large collection of ``unpaired'' video clips (with no fMRI).
Figs.~\ref{decoder nice reconst},~\ref{decoder 1Hz fig} show visual reconstructions of our full \mbox{algorithm (at 0.5Hz and 2Hz, respectively), exhibiting reconstruction of spatial \& temporal details.}


Fig.~\ref{SOTA visual comparison}, compares our 0.5Hz reconstructions to the competing 0.5Hz method of Wen'18~\cite{Wen2018}. 
Our reconstructions capture far more spatial and temporal details.
Fig.~\ref{TSR SOTA visual comparison} compares our HFR results to the competing 4Hz reconstruction method of Wang'22~\cite{Wang2022} (on examples provided in their paper). Our HFR reconstructions are \emph{spatially} more realistic, and much more 
\emph{temporally} consistent across \emph{successive fMRIs} (every 8 frames in Fig.~\ref{TSR SOTA visual comparison} correspond to one fMRI).  \\ \\
\textbf{\underline{\emph{A dynamic video sequence}:}} A live video showing many more video reconstructions can be found on our Project page: 
\href{https://www.wisdom.weizmann.ac.il/~vision/VideoReconstFromFMRI/}{\color{magenta}{\mbox{\small \texttt{https://www.wisdom.weizmann.ac.il/$\sim$vision/VideoReconstFromFMRI/}}}}. The live video shows video reconstruction results for two different framerates (0.5Hz \& 1Hz), from fMRIs of two different subjects (Subjects 1 \& 2 -- the two subjects with highest fMRI SNR; see Appendix B more more details on SNR) .

%
%
%
\begin{table}[b]
\vspace*{-0.5cm}
  \caption{Quantitative evaluation on all reconstruction results}
  \label{tableallsub}
  \centering
  \begin{tabular}{c|ccc|ccc|ccc}
    \toprule    
    {\centering Reconstruction rate}  && Rank$\downarrow$ &&& SSIM$\uparrow$& & &MSE$\downarrow$\\
    \cmidrule(r){1-10}
     & sub1 & sub2 & sub3 & sub1 & sub2 & sub3 & sub1 & sub2 & sub3 \\
    \midrule
    0.5Hz  & 6.84 & 5.86 & 11.8 & 0.135 & \textbf{0.146} & 0.111 & \textbf{0.078} & \textbf{0.078} & 0.092\\
    1Hz & \textbf{5.69} & 4.77 & \textbf{10.2} & \textbf{0.14} & 0.14 & \textbf{0.123} & 0.081 & \textbf{0.078} & 0.094 \\
    2Hz & 5.87 & \textbf{4.1} & 9.88 & 0.129 & 0.141 & 0.119 & 0.08 & 0.081 &  \textbf{0.088} \\
    4Hz &6.97 & 5.81 & 10.07& 0.135 & 0.136 & 0.121 & 0.081& 0.0813& 0.09\\
    \bottomrule
  \end{tabular}
\end{table}
\begin{table}
\centering
\caption{\centering Ablation study of our method (on 0.5Hz Decoder)}
\label{table: ablation super}
\begin{tabular}{llll}
        \toprule
        Model  & Rank $\downarrow$ & SSIM $\uparrow$  & MSE $\downarrow$ \\
        \midrule
        Full method (Internal \& External) &   \textbf{6.84}    & \textbf{0.136} &\textbf{0.078}\\
        Supervised training only &    14.99    & 0.125 &0.084\\
        Self-supervision on Internal only &   7.12    & 0.13 & 0.08\\
        Self-supervision on External only &  11.65    &0.133 & 0.08\\
        \bottomrule
  \vspace*{-0.5cm}
 \end{tabular}
\end{table}
   \hspace{3cm} 
\vspace*{-0.2cm}
%
\subsection{Empirical Evaluation}
\vspace*{-0.1cm}
We numerically compared our reconstruction results to methods that used the same training set.
For these comparisons we used several measures:
\vspace*{-0.25cm}
\begin{itemize}[leftmargin=*]
    \item \textbf{Image similarity measures:}
    We assessed the similarity of individual reconstructed video frames with their ground-truth (GT) counterparts using 
    Structural-Similarity (SSIM) and Mean-Squared Error (MSE). This enabled comparison with the numerical results of~\cite{Wang2022}. 
    \vspace*{-0.15cm}
    \item \textbf{Rank measure:} 
    The above-mentioned standard similarity measures are not very meaningful given the huge visual gap between the GT and reconstructed frames. We therefore used another metric, based on an $n$-way Identification Test.
    We compute the similarity of each short reconstructed clip (e.g., 5 frames) to $n$ short clips extracted from the real test video -- one  is the correct GT clip, and (n-1) distractor clips (i.e., other random clips from the test video).  
    We sort those $n$ computed similarities to obtain a ``Rank score'' -- i.e., the place of the GT clip within this sorted list. Rank=1 is the best score (i.e., the GT clip was most similar), whereas rank=n is the worst. In our experiment, we used $n$=100.
    We used the Perceptual Similarity measure~\cite{Zhang2018} to compare those short clips.
\end{itemize}

Table~\ref{tableallsub} shows our test results for all reconstruction framerates and all subjects. 
Fig.~\ref{SOTA visual comparison} compares our reconstructions to the 0.5Hz method of~\cite{Wen2018}. 
We could numerically compare to them only on the first test clip and only on Subject 1 (as this was their only available reconstruction). Our Decoder \emph{significantly} outperforms~\cite{Wen2018} based on the Wilcoxon rank-sums test
(\underline{Rank:}  w=4647, pvalue=2.82$e^{-56}$; \underline{SSIM:} w=91254, pvalue=2.14$e^{-18}$); 
%
Fig.~\ref{TSR SOTA visual comparison} compares visually and numerically  our HFR reconstructions to the 4Hz method of~\cite{Wang2022} (using their reported numbers). 


    

\vspace*{-0.3cm}
\subsection{Ablation study}
\vspace*{-0.3cm}
We investigate the importance/contribution of specific components of our method. Table~\ref{table: ablation super} shows ablations performed on our 0.5Hz model.
The self-supervised cycle consistency on ``unpaired'' data  significantly improves the reconstruction results compared to supervised training only. This is reflected mostly in the visual results, but also strongly in the numerical Rank measure (Wilcoxon rank-sum test:  \underline{Rank:} w=464652, p=1.5$e^{-41}$; \underline{SSIM:} w=673517, p=0.01; \underline{MSE:} w=2584200, p=8.41$e^{-8}$). 

We further examined the contribution of ``unpaired'' Internal data vs. ``unpaired''  External data in the self-supervised training. As can be see in Table~\ref{table: ablation super}, adding each of them alone provides a significant improvements compared to the supervised-only training.
But using both provides the best results. 

Finally, we compared our HFR reconstruction to simple temporal-interpolation of  0.5Hz reconstruction, to verify that our HFR model indeed recovers new temporal details beyond  interpolation. We found a significant gap in favor of the 1Hz reconstruction on all measures (Wilcoxon rank-sum test: \underline{Rank:} w=1933153, p=1.42$e^{-81}$; \underline{SSIM:} w=2521412, p=3.76$e^{-11}$; \underline{MSE:} w=10872016, p=4$e{-4}$).

%% file: conclusion.tex
\section{Conclusions}
\vspace*{-0.3cm}
We present a self-supervised 
approach for natural-movie reconstruction from fMRI, by imposing cycle-consistency of  2 deep networks: an Encoder, which maps a short video clip to its fMRI response, and a Decoder, which maps an fMRI response to its underlying video clip. Composing these nets sequentially, Encoder-Decoder, allows to train on a large number of "unpaired" video clips (video clips with no corresponding fMRI). This increases the training data by several orders of magnitude, and significantly  improves the reconstruction quality. We further introduced temporal constraints, by learning a self-supervised temporal prior of natural videos, and 
 enforcing temporal continuity on the reconstructed frames. Our self-supervised 
 approach provides state-of-the-art results in natural-video reconstruction from fMRI. It further allows for HFR reconstructions -- decoding fMRI 
 at framerates higher than the underlying fMRI recordings, thus recovering new dynamic details.

%% file: acknowledgments.tex
\section*{Acknowledgments}
This project has received funding from the European Research Council (\textbf{ERC}) under the European Union’s Horizon 2020 research and innovation programme (grant agreement No 788535).

%% file: supplementary.tex
\setcounter{figure}{0}
\setcounter{section}{0}

\renewcommand{\thefigure}{A\arabic{figure}}

\section*{\Large\centering Appendices}


\section*{Appendix A:  Network Architecture}
Figure~\ref{architect fig} below shows the full architecture of the Encoder (Fig.~\ref{architect fig}a) and the 0.5Hz Decoder (Fig.~\ref{architect fig}b). The higher-framerate Decoders are of similar architecture. However, its inputs now are 2 consecutive fMRI recordings, and the outputs are several frames.  Thus, we now learn a transformation from 2 fMRI samples, using the FC layer. We repeat it for each one of the $n$ output frames ans stack the results along new axis, which will serve as the temporal dimension. Then, we use the same architecture as the 0.5Hz architecture (after the first FC layer), which now outputs $\times n$ more frames.  

The second layer of the Encoder architecture is tuned to learn the temporal information using 1D linear combinations of the temporal dimension. It is calculated as follow:
\begin{equation}
    Out_{h,w,c} = \sum_t In_{t,h,w,c}K_t
\end{equation}
where $t,h,w,c$ are the indices of the frame, y location, x location and channel respectively, $In, Out$ are the input and output tensors, and $K$ is the learned kernel. We repeat this process for 8 times, and stack the result along the channel axis, leading to output with $\times 8$ more channels than the input channels.
%

\begin{figure}[h]%

    \centering
    \subfloat{\includegraphics[width=\textwidth]{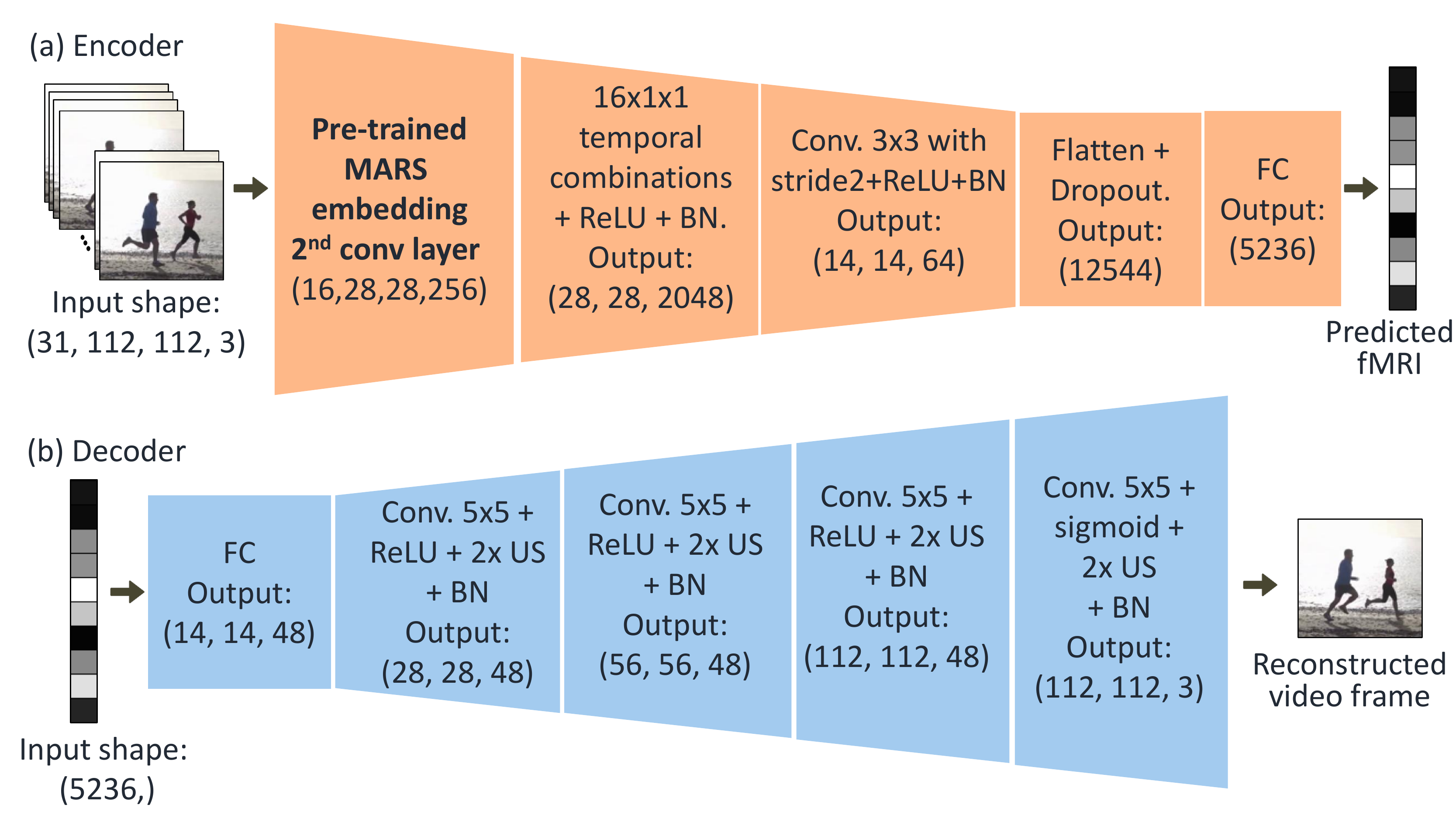}}
    \caption{\textbf{The network architectures.}  {\it (a) Encoder's architecture. (b) Decoder's architecture (shown for the 0.5Hz decoder). ''US'' = Up sample.}
    }
    \label{architect fig}
\end{figure}



\section*{Appendix B: fMRI voxel selection \& SNR estimation} \label{SNR}
The fMRI recordings contain $\sim$100,000 voxels, in the entire brain cortex. Many of these are outside the visual cortex irrelevant for our task. Moreover, many voxels inside the visual cortex are very noisy/distracting. Therefore, for each subject (1,2,3) we separately sift the 5,236 most relevant voxels within each visual cortex -- those voxels with the highest computed signal-to-noise ratio (SNR), as explained below.

The subjects saw each video several times: 2 repetitions for the training videos, and 10 repetitions for the test  videos. 
To select the voxels within the visual cortex only, we followed the scheme of [18]
, which isolates the regions activated by natural movie stimuli (estimated via the intra-subject reproducibility in the fMRI signal across the 2 training repetitions). This results in 10472 detected voxels in the visual cortex.

However, not all voxels in the visual cortex are equally reliable. We sift the half most informative ones. This helps increase reliability of our Encoder and Decoder, as well as reduce the number of parameters in both networks.
%

To account for the differences in voxel reliability, we estimated each voxel's signal to noise ratio (SNR). By 'signal' we refer to the voxel response variance  when  different video frames are presented to the subject. The 'noise' is defined as the voxel response  variance across different repetitions of the same video. The ratio between these 2 measure forms the voxel SNR.
High SNR means that the voxel activation highly changed within the video clips, but was consistent across different repeats of the same clip, indicating that this voxel is informative for the task. \textbf{{Note that estimating the SNR is done directly from the fMRI data, and does not require knowing the ground-truth videos.}}

For each subject we chose the 50\% voxels (out of the 10472 visual cortex voxels) with the highest SNR, leading the final 5236 selected voxels per subject. 

We further estimated \textbf{mean SNR} per subject. The SNRs of the 3 different subjects were:\\
$\bullet$ \ Subject 1: SNR=1.16 \\
$\bullet$ \ Subject 2: SNR=0.92 \\
$\bullet$ \ Subject 3: SNR=0.63. 

As can be seen, the fMRI recording of Subject 3 is of much lower SNR than those of Subjects 1 \& 2 (which was later also reflected in the quality of Subject 3's reconstructions, as evident in Table1 in the paper). Again, note that the quality of a subject's fMRI recordings can be estimated directly from the fMRI data (as explained above), without any reconstruction.


\medskip


%% file: main.bbl
\begin{thebibliography}{19}
\providecommand{\natexlab}[1]{#1}
\providecommand{\url}[1]{\texttt{#1}}
\expandafter\ifx\csname urlstyle\endcsname\relax
  \providecommand{\doi}[1]{doi: #1}\else
  \providecommand{\doi}{doi: \begingroup \urlstyle{rm}\Url}\fi

\bibitem[Beliy et~al.(2019)Beliy, Gaziv, Hoogi, Strappini, Golan, and
  Irani]{Beliy2019}
R.~Beliy, G.~Gaziv, A.~Hoogi, F.~Strappini, T.~Golan, and M.~Irani.
\newblock From voxels to pixels and back: Self-supervision in natural-image
  reconstruction from fmri.
\newblock volume~32, 2019.

\bibitem[Cowen et~al.(2014)Cowen, Chun, and Kuhl]{Cowen2014}
A.~S. Cowen, M.~M. Chun, and B.~A. Kuhl.
\newblock Neural portraits of perception: Reconstructing face images from
  evoked brain activity.
\newblock \emph{NeuroImage}, 94, 2014.
\newblock ISSN 10959572.
\newblock \doi{10.1016/j.neuroimage.2014.03.018}.

\bibitem[Crasto et~al.(2019)Crasto, Weinzaepfel, Alahari, and
  Schmid]{Crasto2019}
N.~Crasto, P.~Weinzaepfel, K.~Alahari, and C.~Schmid.
\newblock Mars: Motion-augmented rgb stream for action recognition.
\newblock volume 2019-June, 2019.
\newblock \doi{10.1109/CVPR.2019.00807}.

\bibitem[Gaziv and Irani(2021)]{Gaziv2021}
G.~Gaziv and M.~Irani.
\newblock More than meets the eye: Self-supervised depth reconstruction from
  brain activity.
\newblock 6 2021.
\newblock URL \url{http://arxiv.org/abs/2106.05113}.

\bibitem[Gaziv et~al.(2022)Gaziv, Beliy, Granot, Hoogi, Strappini, Golan, and
  Irani]{Gaziv2022}
G.~Gaziv, R.~Beliy, N.~Granot, A.~Hoogi, F.~Strappini, T.~Golan, and M.~Irani.
\newblock Self-supervised natural image reconstruction and large-scale semantic
  classification from brain activity.
\newblock \emph{NeuroImage}, page 119121, 7 2022.
\newblock ISSN 10538119.
\newblock \doi{10.1016/j.neuroimage.2022.119121}.

\bibitem[Han et~al.(2019)Han, Wen, Shi, Lu, Zhang, Fu, and Liu]{Han2019}
K.~Han, H.~Wen, J.~Shi, K.~H. Lu, Y.~Zhang, D.~Fu, and Z.~Liu.
\newblock Variational autoencoder: An unsupervised model for encoding and
  decoding fmri activity in visual cortex.
\newblock \emph{NeuroImage}, 198, 2019.
\newblock ISSN 10959572.
\newblock \doi{10.1016/j.neuroimage.2019.05.039}.

\bibitem[Kamitani and Tong(2005)]{Kamitani2005}
Y.~Kamitani and F.~Tong.
\newblock Decoding the visual and subjective contents of the human brain.
\newblock \emph{Nature Neuroscience}, 8, 2005.
\newblock ISSN 10976256.
\newblock \doi{10.1038/nn1444}.

\bibitem[Kay and Gallant(2009)]{Kay2009}
K.~N. Kay and J.~L. Gallant.
\newblock I can see what you see, 2009.
\newblock ISSN 10976256.

\bibitem[Le et~al.(2021)Le, Ambrogioni, Seeliger, Güçlütürk, van Gerven,
  and Güçlü]{Le2021}
L.~Le, L.~Ambrogioni, K.~Seeliger, Y.~Güçlütürk, M.~van Gerven, and
  U.~Güçlü.
\newblock Brain2pix: Fully convolutional naturalistic video reconstruction from
  brain activity.
\newblock \emph{bioRxiv}, 2021.

\bibitem[Naselaris et~al.(2009)Naselaris, Prenger, Kay, Oliver, and
  Gallant]{Naselaris2009}
T.~Naselaris, R.~J. Prenger, K.~N. Kay, M.~Oliver, and J.~L. Gallant.
\newblock Bayesian reconstruction of natural images from human brain activity.
\newblock \emph{Neuron}, 63, 2009.
\newblock ISSN 08966273.
\newblock \doi{10.1016/j.neuron.2009.09.006}.

\bibitem[Nishimoto et~al.(2011)Nishimoto, Vu, Naselaris, Benjamini, Yu, and
  Gallant]{Nishimoto2011}
S.~Nishimoto, A.~T. Vu, T.~Naselaris, Y.~Benjamini, B.~Yu, and J.~L. Gallant.
\newblock Reconstructing visual experiences from brain activity evoked by
  natural movies.
\newblock \emph{Current Biology}, 21, 2011.
\newblock ISSN 09609822.
\newblock \doi{10.1016/j.cub.2011.08.031}.

\bibitem[Qiu et~al.(2017)Qiu, Yao, and Mei]{Qiu2017}
Z.~Qiu, T.~Yao, and T.~Mei.
\newblock Learning spatio-temporal representation with pseudo-3d residual
  networks.
\newblock volume 2017-October, 2017.
\newblock \doi{10.1109/ICCV.2017.590}.

\bibitem[Seeliger et~al.(2018)Seeliger, Güçlü, Ambrogioni, Güçlütürk,
  and van Gerven]{Seeliger2018}
K.~Seeliger, U.~Güçlü, L.~Ambrogioni, Y.~Güçlütürk, and M.~A. van
  Gerven.
\newblock Generative adversarial networks for reconstructing natural images
  from brain activity.
\newblock \emph{NeuroImage}, 181, 2018.
\newblock ISSN 10959572.
\newblock \doi{10.1016/j.neuroimage.2018.07.043}.

\bibitem[Shen et~al.(2019)Shen, Dwivedi, Majima, Horikawa, and
  Kamitani]{Shen2019}
G.~Shen, K.~Dwivedi, K.~Majima, T.~Horikawa, and Y.~Kamitani.
\newblock End-to-end deep image reconstruction from human brain activity.
\newblock \emph{Frontiers in Computational Neuroscience}, 13, 2019.
\newblock ISSN 16625188.
\newblock \doi{10.3389/fncom.2019.00021}.

\bibitem[Simonyan and Zisserman(2015)]{Simonyan2015}
K.~Simonyan and A.~Zisserman.
\newblock Very deep convolutional networks for large-scale image recognition.
\newblock 2015.

\bibitem[Tran et~al.(2014)Tran, Bourdev, Fergus, Torresani, and
  Paluri]{Tran2014}
D.~Tran, L.~Bourdev, R.~Fergus, L.~Torresani, and M.~Paluri.
\newblock Learning spatiotemporal features with 3d convolutional networks.
\newblock 12 2014.
\newblock URL \url{http://arxiv.org/abs/1412.0767}.

\bibitem[Wang et~al.(2022)Wang, Yan, Huang, Li, Wang, Fan, Sheng, Liu, Li, and
  Chen]{Wang2022}
C.~Wang, H.~Yan, W.~Huang, J.~Li, Y.~Wang, Y.-S. Fan, W.~Sheng, T.~Liu, R.~Li,
  and H.~Chen.
\newblock Reconstructing rapid natural vision with fmri-conditional video
  generative adversarial network.
\newblock \emph{Cerebral Cortex}, 1 2022.
\newblock ISSN 1047-3211.
\newblock \doi{10.1093/cercor/bhab498}.

\bibitem[Wen et~al.(2018)Wen, Shi, Zhang, Lu, Cao, and Liu]{Wen2018}
H.~Wen, J.~Shi, Y.~Zhang, K.~H. Lu, J.~Cao, and Z.~Liu.
\newblock Neural encoding and decoding with deep learning for dynamic natural
  vision.
\newblock \emph{Cerebral Cortex}, 28, 2018.
\newblock ISSN 14602199.
\newblock \doi{10.1093/cercor/bhx268}.

\bibitem[Zhang et~al.(2018)Zhang, Isola, Efros, Shechtman, and Wang]{Zhang2018}
R.~Zhang, P.~Isola, A.~A. Efros, E.~Shechtman, and O.~Wang.
\newblock The unreasonable effectiveness of deep features as a perceptual
  metric.
\newblock 2018.
\newblock \doi{10.1109/CVPR.2018.00068}.

\end{thebibliography}
